# Bidirectional Encoder Representations from Transformers (BERT): A sentiment analysis odyssey

Shivaji Alaparthi[1] and Manit Mishra[2]

The purpose of the study is to investigate the relative effectiveness of four different sentiment analysis techniques: (1) unsupervised lexicon-based model using Sent WordNet; (2) traditional supervised machine learning model using logistic regression; (3) supervised deep learning model using Long Short-Term Memory (LSTM); and, (4) advanced supervised deep learning models using Bidirectional Encoder Representations from Transformers (BERT). We use publicly available labeled corpora of 50,000 movie reviews originally posted on internet movie database (IMDB) for analysis using Sent WordNet lexicon, logistic regression, LSTM, and BERT. The first three models were run on CPU based system whereas BERT was run on GPU based system. The sentiment classification performance was evaluated based on accuracy, precision, recall, and F1 score. The study puts forth two key insights: (1) relative efficacy of four highly advanced and widely used sentiment analysis techniques; (2) undisputed superiority of pre-trained advanced supervised deep learning BERT model in sentiment analysis from text data. This study provides professionals in analytics industry and academicians working on text analysis key insight regarding comparative classification performance evaluation of key sentiment analysis techniques, including the recently developed BERT. This is the first research endeavor to compare the advanced pre-trained supervised deep learning model of BERT vis-à-vis other sentiment analysis models of LSTM, logistic regression, and Sent WordNet.

(1) Data Scientist, CenturyLink, Bengaluru, India, E-mail id: *alaparthishivaji@gmail.com*

(2) Associate Professor, International Management Institute Bhubaneswar, India, E-mail id: *manit.mishra@imibh.edu.in*

**Introduction**

Data is the new oil (Yi *et al.*, 2014), the most sought after raw material in 21$^{st}$ century (Berners-Lee and Shadbolt, 2011). It is a bottomless container of insight for organizations as every single day 2.5 quintillion bytes (2.5X10$^{18}$ bytes) of data gets added (Dobre and Xhafa, 2014). Such an inflow is inevitable given the fact that only Walmart collects about 2.5 petabytes (2.5X10$^{15}$) of data regarding customer transactions every hour (McAfee *et al.*, 2012). The data deluge is only going to be more pronounced in future with continuous emergence of new data sources (Bradlow *et al.*, 2017). A significant portion of the data being generated is text data – structured, semi-structured, and unstructured (Sivarajah *et al.*, 2017). Online consumer reviews are a formidable source of such text data which have been utilized in extant research for the purpose of predicting consumer ratings (Buschken and Allenby, 2016); forecasting stock market performance (Tirunillai and Tellis, 2012); extracting underlying dimensions of consumer satisfaction with quality (Tirunillai and Tellis, 2014); analysis and visualization of market structure (Lee and Bradlow, 2011); generating market structure and competitive landscape insights (Netzer *et al.*, 2012); and even for assessing box office performance of movies (Chintagunta *et al.*, 2010). Wedel and Kannan (2016, p. 107) underscore the significance of text data for future research in the area of marketing when they emphasize that "models for largescale unstructured data are still in their infancy, but research is starting to emerge."

The opportunities notwithstanding, the challenges in deriving meaning out of text data are also manifold. First, it is required to derive meaning from word or sentence in the context in which it has been used (Buschken and Allenby, 2016). For example, raging bull has a different meaning in the review of stock market as compared to a narrative about tourist attractions in Spain. Second, the review itself may be fuzzy and/or incoherent thereby making it difficult to infer its meaning (Archak *et al.*, 2011). For example, a movie review might read, "lackluster script saved by star power." Such a dichotomous statement puts a question mark on its valence. Third, unlike survey data, text data is not readily available for data analysis. It requires substantial preprocessing before it can be subjected to statistical analysis (Tirunillai and Tellis, 2014). An exhaustive description of the various steps involved in preprocessing of text prior to its statistical can be found in Bird *et al.*'s (2009) Natural Language Tool Kit (NLTK) book. Given the enormity of text mining task, over a period of time multiple algorithms have been made available by programming enthusiast, data scientists, and even leading corporate houses such as Amazon.

The intent of this paper is to offer a comparative analysis of the various algorithm options available for an important application of text analytics - sentiment analysis. Sentiment analysis has been cited as one of the most popular applications of text analytics (Sarkar, 2019). Pang and Lee (2008, p. 10), in one of the early attempts to summarize research on consumer sentiments, have defined sentiment analysis as an "automatic analysis of evaluative text and tracking of the predictive judgments therein." A more elaborate definition has been given by Liu and Zhang (2012, p. 415) who define sentiment analysis as "the computational study of people's opinions, appraisals, attitudes, and emotions toward entities, individuals, issues, events, topics, and their attributes." Assessment of stakeholder sentiments has always been an important parameter for an organization's sustenance and excellence. It is in complete alignment with the very foundational concepts of marketing such as market orientation (Kohli and Jaworski, 1990). The significance of sentiment analysis has increased manifold since the availability of social media platforms to vent opinions (Liu and Zhang, 2012). However, keeping pace with the explosion in availability of text reflecting sentiments, multiple tools and techniques have inundated the analytics ecosystem. Therefore, in order to offer analytics enthusiasts a one-stop source on relative efficacy of some of the prominent techniques of sentiment analysis, this study derives sentiments from a large corpus of publicly available movie reviews using four of the more widely used techniques: (1) unsupervised lexicon-based models e.g., Sent WordNet; (2) traditional supervised machine learning models e.g., logistic regression; (3) supervised deep learning models e.g., Long Short-Term Memory (LSTM); and, (4) advanced supervised deep learning models e.g., Bidirectional Encoder Representations from Transformers (BERT). Based on sentiment analysis of this corpus of big data, we present output quality ratings of the above-mentioned techniques of assessing sentiments from text. To the best of our knowledge, no single comprehensive repository of information exists today which covers all the techniques mentioned above. Therefore, the paper would serve as a source of reference for relative effectiveness of some of the most advanced techniques of conducting sentiment analysis.

**Sentiment analysis techniques: A review**

Consumer sentiments are a precursor to consumer intentions and decisions. The sentiments are hidden in the text uploaded onto various social media platforms by consumers in the form of reviews. This unstructured text data can be analyzed using sentiment analysis techniques to derive a qualitative as well as quantitative interpretation of the underlying sentiments. While a qualitative interpretation entails assessment on a positive to negative scale, a quantitative interpretation facilitates calculation of sentiment polarity and subjectivity vis-à-vis objectivity proportions. The primary target of analysis is the body of text comprising of all reviews called as corpora. The corpora is made of a large number of individual text sentences or paragraphs called as documents. It is the analysis of corpora made up of documents that leads to derivation of sentiments. Over a period of time there has been a gradual increase in the number and types of approaches for undertaking sentiment analysis.

*Preprocessing text data*

Text data is largely unstructured (Sivarajah *et al.*, 2017) and therefore, requires substantial preprocessing before sentiment analysis can be carried out (Tirunillai and Tellis, 2014). There are a number of good sources to understand preprocessing (e.g., Aggarwal and Zhai, 2012; Sarkar, 2019) so we would only touch upon the major steps before moving on to the objective of this study – comparison of sentiment analysis techniques. The pre-processing of text data prior to conducting sentiment analysis involves following steps:

1. Removing HTML tags from the text e.g. "<"
2. Removing accented characters e.g. "ύ"
3. Expanding the contracted words e.g. converting "haven't" to "have not"
4. Removing special characters e.g. "@" and "#"
5. Separating text into sentences based on presence of punctuations and thereafter, removing the punctuations e.g. "," or "."
6. Lemmatization to arrive at the root word e.g. the root word for "succeeding" and "successful" would be "success"
7. Replacing capital letters with lower-case letters e.g. "Positive" replaced by "positive"
8. Removing rare words that appear in less than 1% of the documents in any corpus
9. Removing stop words that do not contribute to the meaning e.g. "the"

The reviews that form part of the corpus are generated by individuals from different cultures having different ways of communicating in the same language and with no incentive to be grammatically correct. Therefore, preprocessing of the text is essential to bring some degree of standardization to the corpus and make it amenable to further analysis.

*Unsupervised lexicon-based model: Sent WordNet*

These set of techniques are unsupervised and based on a lexicon or dictionary created and curated specifically for sentiment analysis. The lexicons contain information on words; related positive or negative sentiment; polarity as described by magnitude of positivity or negativity in the words; parts of speech (POS) tag; subjectivity classification in terms of strong, weak, or neutral etc. Sentiment analysis using this technique involves associating the words in the corpus with the corresponding information provided in the lexicon along with contextual information in order to derive sentiment scores of documents constituting the corpora. Some of the popular lexicons in use are TextBlob, AFINN, VADER, and Sent WordNet.

*Supervised machine learning model: Logistic regression*

The traditional supervised machine learning predictive algorithm can also be used to train the model into classifying sentiments. The sentiments undergo a classification into positive or negative based on the review content. This is followed by model validation. One of the more popular technique for undertaking such a binary classification is logistic regression (logit). The logit model is based on the theory of utility maximization incorporating a random factor (Nijkamp et al., 1992). Sentiment analysis using logistic regression is based on bag of words model wherein each document in the corpora is considered to be a bag of unstructured words irrespective of its context, order, or grammar. This results is a term-document matrix wherein each document is a row and each unique word is a distinct column. Thus, a corpus comprising of *d* documents (reviews) and *t* unique terms (words) results in a (*d* X *t*) matrix which is subjected to logistic regression. The logit model calculates the odds of a document having a positive sentiment and based on its comparison with a cut-off value, is classified as either a positive or a negative sentiment. The logit model uses a sigmoidal function and takes values between -∞ to +∞. It is a useful supervised machine learning algorithm because it doesn't have to assume linearity.

*Supervised deep learning model: LSTM*

The supervised deep learning models based on neural network architecture are being increasingly used to solve complicated problems and provide intelligent solutions (Metaxiotis and Psarras, 2004). Its nonparametric nature, capability to train and build complex non-linear models, and its ability to handle missing data makes it a sought after technique (Venugopal and Baets, 1994). The most widely used neural network algorithm is the one based on feed forward mechanism and backward propagation of error which facilitates analysis over input, output, and multiple hidden layers (Shmueli *et al.*, 2007). A variant of the deep neural network is the long short-term memory (LSTM) displaying a recurrent neural network (RNN) architecture which has applications in sentiment analysis of text data. In recent times, the standard LSTM architecture networks "have become the state-of-the-art models for a variety of machine learning problems" (Greff *et al.*, 2016, p. 2222). LSTM units generally comprise of a cell, an input gate, an output gate, and a forget gate. LSTMs are bidirectional networks that have been found to be more accurate in comparison to the traditional multilayer perceptrons as well as standard RNNs (Graves and Schmidhuber, 2005).

*Advanced supervised deep learning model: BERT*

Sentiment analysis from text data has undergone a colossal transformation with the arrival of pre-trained transformer models such as Bidirectional Encoder Representations from Transformers (BERT). Developed by Devlin *et al.* (2018) of Google AI Language, BERT is "designed to pretrain deep bidirectional representations from unlabeled text by jointly conditioning on both left and right context in all layers." The state-of-the-art BERT is pre-trained on two unsupervised tasks – masked language modeling and next sentence prediction, thus making it an effective technique for sentiment classification (Trivedi 2019, January 27). BERT is an advanced as well as more realistic technique since it accepts the fact that a document can simultaneously belong to multiple classes. BERT is known to have achieved exceptional results in eleven natural language understanding (NLU) tasks (Devlin *et al.*, 2018). The credibility of BERT can be inferred from the fact that Google uses it in its search algorithms (Nayak, 2019) and is currently applicable to over 70 different languages (Roger, 2019).

**Analysis and interpretation**

*Data*

The study uses publicly available movie review data from internet movie database (IMDB). The dataset was first introduced and curated by Maas *et al.* (2011) in their study on vector-based approaches to sentiment classification. This dataset consists of 50,000 movie reviews such that each movie accounts for 30 or lesser number of reviews. Each review has been labeled *a priori* as having either positive or negative sentiment. There are an equal number of positive and negative sentiment reviews in the dataset. The reviews have a sentiment polarity score of either $\leq 4$ or $\geq 7$ on a polarity index of 10. There are no neutral reviews in the dataset. The dataset was downloaded from http://ai.stanford.edu/~amaas/data/sentiment/.

*Analysis*

For the purpose of this study, the outputs for the techniques unsupervised lexicon based model using Sent WordNet, unsupervised machine learning model using logistic regression, and supervised deep learning model using LSTM, were derived based on a partitioning of 60% training (30,000 reviews) and 40% validation (20,000 reviews) data. However, for the advanced supervised deep learning model using BERT, partitioning was done into 35% training (17,500 reviews), 15% validation (7,500 reviews), and 50% test (25,000 reviews) data. The structure of partitioning for BERT was altered due to two reasons: (1) possibility of overfitting since BERT is a pre-trained model; and, (2) requirement of greater computational power for a larger training data. The classification performance of the models is assessed based on their accuracy, precision, recall, and F1 score obtained for sentiment classification of validation data or test data, in case of BERT.

The accuracy or hit-ratio of the model is the total number of documents correctly classified. It provides the overall predictive accuracy of the model in classifying the reviews into positive and negative sentiments. The precision of the model is determined from the ratio of correctly classified documents out of all the documents that have been predicted as having a positive sentiment. It suggests success in classification within the predicted reviews belonging to important class – reviews having a positive sentiment. The recall or sensitivity is the true positive ratio. It indicates the proportion of actual documents belonging to the important class that have been correctly classified. Recall is a measure of the robustness of the model because it reflects the ability of the model to correctly predict the reviews having a positive sentiment. While recall gives the ability of the model to correctly predict the documents that actually belong to important class, precision portrays the model's success rate out of all documents predicted to belong to the important class – in our case, the positive sentiment class. The F1 score of the model provides a balance between the two and is an amalgamation of both these measures. A high F1 score indicates a model that not only has high predictive ability of the important class but also has high success rate within the predictions of the important class. For all four measures of model robustness, a value closer to 1 is considered as better. The F1 score is obtained using the formula:

$$F1\ score = 2 * ((precision * recall)/(precision + recall))$$

The classification performance of all the four techniques on the four output measures was derived based on their ability to classify the sentiments expressed in the reviews in validation data.

The sentiment analysis of the corpora based on Sent WordNet, logistic regression, and LSTM was carried out on a central processing unit (CPU) based system whereas BERT was executed on a graphics processing unit (GPU) based system. The run time using BERT for 5 epochs was 100 minutes. The derived classification performance outputs are given in Table 1.

**Table 1.** Classification performance outputs

| Classification technique | Model | Accuracy | Precision | Recall | F1 Score |
|---|---|---|---|---|---|
| Unsupervised lexicon based model | Sent WordNet | 0.6308 | 0.6747 | 0.6308 | 0.6064 |
| Supervised machine learning model | Logistic regression | 0.8941 | 0.8975 | 0.8941 | 0.8941 |
| Supervised deep learning model | LSTM | 0.8675 | 0.8680 | 0.8675 | 0.8675 |
| Advanced supervised deep learning model | BERT | 0.9231 | 0.9235 | 0.9231 | 0.9231 |

**Discussion and conclusion**

The study examined the different sentiment analysis techniques on a publicly available labeled dataset of 50,000 IMDB movie reviews. Sentiment classification was carried out using unsupervised lexicon based model using Sent WordNet, supervised machine learning model using logistic regression, supervised deep learning model using LSTM, and advanced supervised deep learning model using pre-trained BERT. A comparative analysis of the four models reveal undisputed superiority of the pre-trained BERT model in sentiment classification. BERT is capable of achieving "deep bidirectional representations from unlabeled text by jointly conditioning on both left and right context in all layers" (Devlin *et al.*, 2018). This is unlike other language representation models. That explains its unmatched superiority in sentiment classification in our study. Our findings also stand substantiated by BERT's 80.5% score on general language understanding evaluation (GLUE) benchmark (Wang *et al.*, 2018). Poor performance of LSTM in comparison to logistic regression could be due to the fact that LSTM's key strength lies in dealing with vanishing gradient problems (Sundermeyer *et al.*, 2012). Therefore, it is more suitable for cases having a sequence of data points e.g., videos rather than single data points e.g., text or images. The performance of unsupervised lexicon based model using the Sent WordNet lexicon is comparatively inferior to other models probably due to its unsupervised predictive algorithm.

The study sheds new light on comparative classification performance evaluation of various sentiment analysis techniques on a labeled corpora. This is meant to aid analytics professionals pursuing sentiment analysis. To the best of our knowledge, this is the first research endeavor to compare the advanced pre-trained supervised deep learning model of BERT vis-à-vis other models. Even as the study broadens the horizon of work on sentiment analysis techniques, a few limitations need to taken note of. First, BERT demands strong computational capabilities. A greater computational power could have allowed us to train the model on a larger data with more epochs leading to an even stronger performance by the model. Second, sentiment analysis using both labeled as well unlabeled data could have thrown interesting results. Having said that, we have endeavored to provide a platform for future studies on sentiment analysis model comparison studies. The insight generated study can be used by academicians and industry experts executing sentiment analysis for improved sentiment classification using a proven superior technique.